\documentclass[acmtog,nonacm]{acmart}

\AtBeginDocument{%
  \providecommand\BibTeX{{%
    \normalfont B\kern-0.5em{\scshape i\kern-0.25em b}\kern-0.8em\TeX}}}

\acmYear{2025} 

\usepackage[ruled]{algorithm2e} 

\SetAlFnt{\small}
\SetAlCapFnt{\small}
\SetAlCapNameFnt{\small}
\SetAlCapHSkip{0pt}

\usepackage{float} 
\usepackage{graphicx}
\usepackage{amsmath}
\usepackage{booktabs}
\usepackage{multirow}
\usepackage{makecell} 
\usepackage{colortbl}
\usepackage{threeparttable}
\usepackage[misc]{ifsym}
\usepackage{pifont}
\usepackage[misc]{ifsym}
\usepackage{color}

\usepackage{enumitem}
\usepackage{subcaption}
\usepackage{xspace}

\usepackage[capitalize]{cleveref}
\crefname{section}{Sec.}{Secs.}
\Crefname{section}{Section}{Sections}
\Crefname{table}{Table}{Tables}
\crefname{table}{Tab.}{Tabs.}

\newcommand{\name}{\textsc{LayerPano3D}}
\newcommand{\dataname}{\textsc{Upright360~}}

\begin{document}

\title{\name: Layered 3D Panorama for Hyper-Immersive Scene Generation}

\author{Shuai Yang}
\authornotemark[1]

\affiliation{
    \institution{Shanghai Jiao Tong University}
    \country{China}
}
\affiliation{%
    \institution{Shanghai AI Laboratory}
	\country{China}
}

\author{Jing Tan}
\authornotemark[1]

\affiliation{%
    \institution{The Chinese University of Hong Kong}
	\country{China}
}
\affiliation{%
    \institution{Shanghai AI Laboratory}
	\country{China}
}

\author{Mengchen Zhang}
\affiliation{%
    \institution{Zhejiang University}
	\country{China}
}
\affiliation{%
    \institution{Shanghai AI Laboratory}
	\country{China}
}

\author{Tong Wu}
\authornotemark[2]
\affiliation{%
    \institution{The Chinese University of Hong Kong}
	\country{China}
}
\affiliation{%
    \institution{Shanghai AI Laboratory}
	\country{China}
}

\author{Yixuan Li}
\affiliation{%
    \institution{The Chinese University of Hong Kong}
	\country{China}
}
\affiliation{%
    \institution{Shanghai AI Laboratory}
	\country{China}
}

\author{Gordon Wetzstein}
\affiliation{%
    \institution{Stanford University}
	\country{USA}
}

\author{Ziwei Liu}
\affiliation{%
    \institution{Nanyang Technological University}
	\country{Singapore}
}
\author{Dahua Lin}
\authornotemark[2]
\affiliation{%
    \institution{The Chinese University of Hong Kong}
	\country{China}
}
\affiliation{%
    \institution{Shanghai AI Laboratory}
	\country{China}
}

\makeatletter
\let\@authorsaddresses\@empty
\makeatother
\begin{abstract}
3D immersive scene generation is a challenging yet critical task in computer vision and graphics. 
A desired virtual 3D scene should 1) exhibit omnidirectional view consistency, and 2) allow for large-range exploration in complex scene hierarchies. 
Existing methods either rely on successive scene expansion via inpainting or employ panorama representation to represent large FOV scene environments. However, the generated scene suffers from semantic drift during expansion and is unable to handle occlusion among scene hierarchies.
To tackle these challenges, we introduce\textbf{~\name}, a novel framework for full-view, explorable panoramic 3D scene generation from a single text prompt. 
Our key insight is to decompose a reference 2D panorama into multiple layers at different depth levels, where each layer reveals the unseen space from the reference views via diffusion prior.~\name~comprises multiple dedicated designs: 
\textbf{1)} We introduce a new panorama dataset \dataname, comprising 9k high-quality and upright panorama images, and finetune the advanced Flux model on \dataname for high-quality, upright and consistent panorama generation related tasks.
\textbf{2)} We pioneer the Layered 3D Panorama as underlying representation to manage complex scene hierarchies and lift it into 3D Gaussians to splat detailed 360-degree omnidirectional scenes with unconstrained viewing paths.
Extensive experiments demonstrate that our framework generates state-of-the-art 3D panoramic scene in both full view consistency and immersive exploratory experience. We believe that~\name~holds promise for advancing 3D panoramic scene creation with numerous applications. More examples please visit our project page: \href{https://layerpano3d-web.github.io}{\color{purple}ys-imtech.github.io/projects/LayerPano3D/}

\end{abstract}

\begin{teaserfigure}
\centering
    \includegraphics[width=\linewidth]{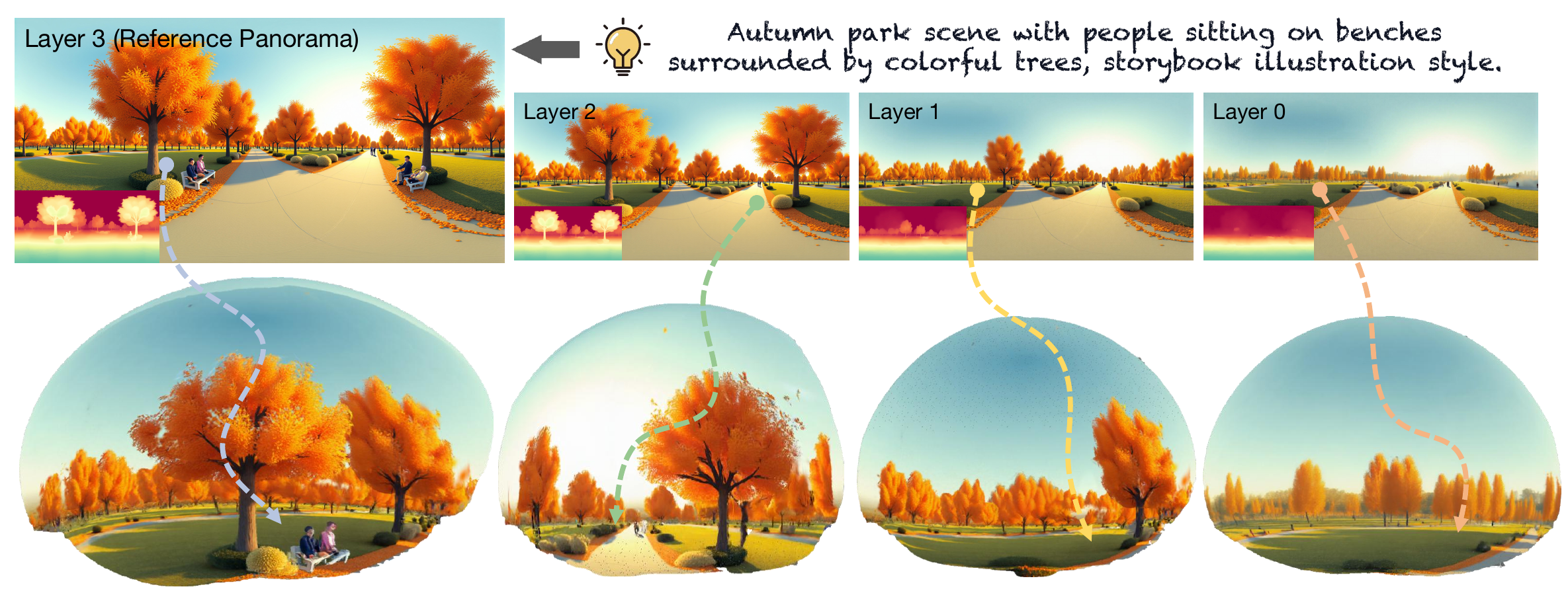}
    \caption{\small \textbf{Overview of~\name.} Guided by simple text prompts,~\name~leverages multi-layered 3D panorama to create hyper-immersive panoramic scene with $360^\circ \times 180^\circ$ coverage, enabling free 3D exploration among complex scene hierarchies. }
  \label{fig:teaser}
\end{teaserfigure}

\maketitle

\section{Introduction}
The development of spatial computing, including virtual and mixed reality systems, greatly enhances user engagement across various applications, and drives demand for explorable, high-quality 3D environments. We contend that a desired virtual 3D scene should 1) exhibit high-quality and consistency in appearance and geometry across the full $360^\circ \times 180^\circ$ view; 2) allow for exploration among complex scene hierarchies with clear parallax. In recent years, many approaches in 3D scene generation~\cite{gao2024cat3d,li2024dreamscene,zhang2023scenewiz3d} were proposed to address these needs.

One branch of works~\cite{luciddreamer, wonderjourney, text2room, fridman2023scenescape, ouyang2023text2immersion} seeks to create extensive scenes by leveraging a ``navigate-and-imagine'' strategy, which successively applies novel-view rendering and outpaints unseen areas to expand the scene. However, this type of approaches suffer from the semantic drift issue: long sequential scene expansion easily produces incoherent results as the out-paint artifacts accumulate through iterations, hampering the global consistency and harmony of the generated scene.  

Another branch of methods~\cite{tang2023mvdiffusion, zhang2024taming, wang2022stylelight, wang2023customizing, chen2022text2light} employs Equirectangular Panorama to represent $360^\circ$, large field of view (FOV) environments in 2D. 
However, the absence of large-scale panoramic datasets hinders the capability of panorama generation systems, resulting in low-resolution images with simple structures and sparse assets.
{Moreover, 2D panorama~\cite{tang2023mvdiffusion,zhang2024taming,wang2022stylelight} does not allow for flexible scene exploration. Even when lifted to a panoramic scene~\cite{zhou2024dreamscene360}, the simple spherical structure fails to provide complex scene hierarchies with clear parallax, leading to occluded spaces that cause blurry renderings, ambiguity, and gaps in the generated 3D panorama. Some methods~\cite{zhou2024holodreamer} typically use inpainting-based disocclusion strategy to fill in the unseen spaces, but they require specific, predefined rendering paths tailored for each scene, limiting the potential for flexible exploration. }

To this end, we present \name, a novel framework that leverages Multi-Layered 3D Panorama for explorable, full-view consistent scene generation from text prompts. The main idea is to create a Layered 3D Panorama by first generating a reference panorama and treating it as a multi-layered composition, where each layer depicts scene content at a specific depth level. In this regard, it allows us to create complex scene hierarchies by placing occluded assets in different depth layers at full appearance. 

Our contributions are two-fold. 
\textbf{First}, to generate high-quality and coherent $360^\circ \times 180^\circ$ panoramas, 
we curate a new dataset, namely \dataname, consisting of 9k high-quality, upright panorama images, and finetune the advanced Flux~\cite{flux2023} model with panorama LoRA on it for panorama generation and inpainting. This feed-forward pipeline prevents semantic drifts during panorama generation, while ensuring a consistent horizon level across all views.

{\textbf{Second}, we introduce the Layered 3D Panorama representation as a general solution to handle occlusion for different types of scenes with complex scene hierarchies, and lift it to 3D Gaussians~\cite{3dgs} to enable large-range 3D exploration. By leveraging pre-trained panoptic segmentation prior and K-Means clustering, we streamline an automatic layer construction pipeline to decompose the reference panorama into different depth layers. The unseen space at each layer is synthesized with the Flux-based inpainting pipeline.  

Extensive experiments demonstrate the effectiveness of~\name~in generating hyper-immersive layered panoramic scene from a single text prompt.
\name~surpasses state-of-the-art methods in creating coherent, plausible, text-aligned 2D panorama and full-view consistent, explorable 3D panoramic environments.
Furthermore, our framework does not require any scene-specific navigation paths, 
providing more user-friendly interface for non-experts.
We believe that~\name~effectively enhances the accessibility of full-view, explorable AIGC 3D environments for real-world applications.
\section{Related Works}

\subsection{3D Scene Generation}
Due to the recent success of diffusion models~\cite{dreamgaussian, poole2022dreamfusion}, 3D scene generation has also achieved some development. Scenescape~\cite{fridman2023scenescape} and DiffDreamer~\cite{DBLP:conf/iccv/CaiCPSOGW23}, for example, explore perpetual view generation through the incremental construction of 3D scenes. One major branch of work employ step-by-step inpainting from pre-defined trajectories. Text2Room~\cite{text2room} creates room-scale 3D scenes based on text prompt, utilizing textured 3D meshes for scene representation. Similarly, LucidDreamer~\cite{luciddreamer} and WonderJourney~\cite{wonderjourney} can generate domain-free 3D Gaussian splatting scenes from iterative inpainting. However, this line of work often suffer from the semantic drift issue, resulting in unrealistic scene from artifact accumulation and inconsistent semantics. While some other approaches~\cite{cohenbar2023setthescene, zhang2023scenewiz3d, vilesov2023cg3d} endeavor to integrate objects with environments, they yield relatively low quality of comprehensive scene generation. 
Recently, our concurrent works, DreamScene360~\cite{zhou2024dreamscene360} and HoloDreamer~\cite{zhou2024holodreamer} also employ panorama as prior to construct panoramic scenes. However, they only achieve the $360^\circ \times 180^\circ$ field of view at a fixed viewpoint based on a single panorama of low-quality and simple structure, and do not support free roaming within the scene. In contrast, our framework leverages Multi-Layered 3D Panorama representation to construct high-quality, fully enclosed scenes that enable larger-range exploration paths in 3D scene.

\vspace{-9pt}

\subsection{Panorama Generation}
Panorama generation methods are often based on GANs or diffusion models. Early in this field, with the different forms of deep generative neural networks, GAN-based panorama generation methods explore many paths to improve quality and diversity. Among them, Text2Light~\cite{chen2022text2light} focuses on HDR panoramic images by employing a text-conditioned global sampler alongside a structure-aware local sampler. However, training GANs is challenging and they encounter the issue of mode collapse. Recently, some studies have utilized diffusion models to generate panoramas. MVDiffusion~\cite{tang2023mvdiffusion} generates eight perspective views with multi-branch UNet but the resulting closed-loop panorama only captures the $360^\circ \times 90^\circ$ FOV. The image generated from MultiDiffusion~\cite{bartal2023multidiffusion} and Syncdiffusion~\cite{lee2023syncdiffusion} is more like a long-range image with wide horizontal angle as they do not integrate camera projection models. PanoDiff~\cite{Wangpanodiff} can generate $360^\circ$ panorama from one or more unregistered Narrow Field-of-View (NFoV) images with pose estimation and controlling partial FOV LDM, while the quality and diversity of results are limited by the scarcity of panoramic image training data like most other methods~\cite{wang2023customizing, li2023panogen, wu2024panodiffusion}. In contrast, our model can generate Multi-Layered 3D Panorama for immersive, high-quality, and coherent scene generation from text prompts.

\section{Method}
\begin{figure*}[t]
	\centering
	\includegraphics[width=1.0\linewidth]{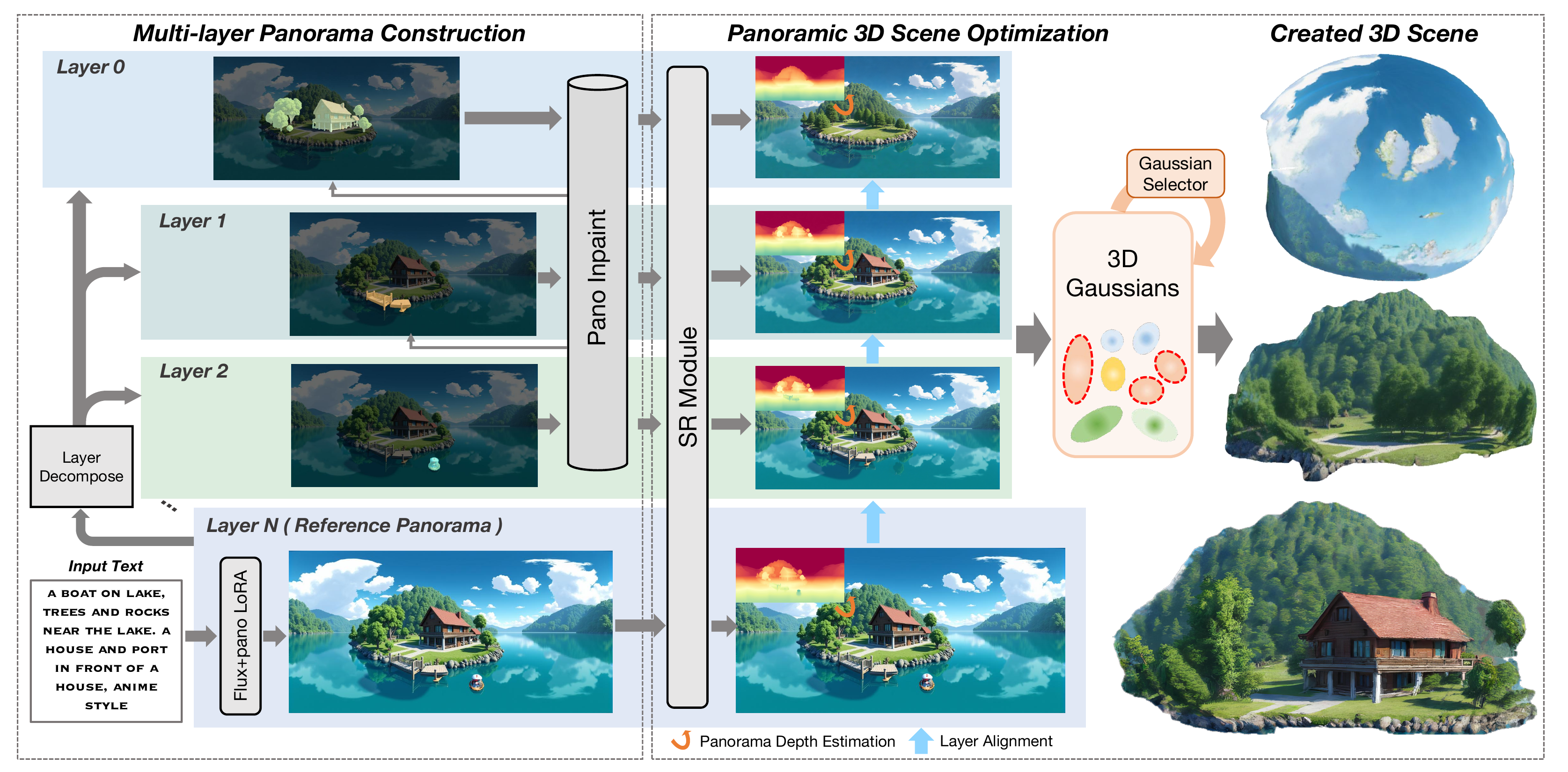}
        \setlength{\abovecaptionskip}{-3mm}
	\caption{\small
	\textbf{Pipeline Overview of~\name.} Our framework consists of two stages, namely multi-layer panorama construction and panoramic 3D scene optimization. ~\name~streamlines an automatic generation pipeline without any manual efforts to design scene-specific navigation paths for expansion or completion.
	}
	\label{fig:pipeline}
    \vspace{-10pt}
\end{figure*}
The goal of our work is to create a panoramic scene guided by text prompts. This generated scene encompasses a complete $360^\circ \times 180^\circ$ field of view from various viewpoints within an extensive range in the scene, while allowing for immersive exploration along complex trajectories. \name~consists two stages. In \textbf{Stage I} (\cref{sec:method1}), we first generate the reference panorama from text prompt by finetuning flux~\cite{flux2023} with panorama LoRA in our \dataname dataset. With the reference panorama, we construct our Layered 3D Panorama representation by iterative layer decomposition, completion and alignment process. In \textbf{Stage II} (\cref{sec:method2}), the Layered 3D Panorama is lifted to 3D Gaussians in a cascaded manner to enable large-range 3D exploration.

\subsection{Multi-Layer Panorama Generation}
\label{sec:method1}
We introduce the Layered 3D Panorama representation based on the following assumption: ``\textit{an enclosed 3D scene contains a background and various assets positioned in front of it}''. In this regard, using Layered 3D Panorama for 3D scene generation is a general approach to handle occlusion for various types of scene. To create a complete scene, we first generate a high-quality reference panorama from a single text prompt and decompose it into $N+1$ layers along the depth dimension. As shown in~\cref{fig:pipeline}, these layers, arranged from the farthest to the nearest, represent both the scene background (layer 0) and the layouts behind the observation point.

\subsubsection{Reference Panorama Generation and \dataname.}
A good panorama captures rich scene details, creating an immersive and comprehensive view. This depth of detail fosters a stronger connection and deeper comprehension of the scene. Unlike the advancement in standard image generation, panorama generation still faces quality gaps. To ensure upright and geometrically consistent scenes, we curate a new dataset, namely \dataname, comprising high quality, upright panorama images, and finetune the Flux~\cite{flux2023} with LoRA on the \dataname dataset. This lightweight training approach achieves optimal performance even with limited high-quality panorama data and can be directly extended to subsequent tasks, for example panorama inpainting for layer completion. For data curation, we first collect around 15k raw panorama images: 9684 panorama image samples are collected from Matterport3D~\cite{Matterport3D}, 1824 images from the web, and 3592 synthetic panoramas generated from Blockadelabs. Based on this, we leverage GeoCalib~\cite{GeoCalib}, a state-of-the-art single-perspective-image calibration method, to filter upright panoramas. Specifically, we employ the Equirectangular Projection (ERP), a mapping technique that projects a 3D sphere onto a 2D plane, to generate four perspective views from each panorama. These views are extracted at a fixed field of view (FOV) of 90°, an elevation of 0°, and four distinct azimuths (0°, 90°, 180°, 270°). We then use GeoCalib to calibrate the four views, computing their pitch and roll variances for filtering. Panoramas with variances exceeding 1.0 are classified as non-upright and excluded: $Var({pitch_{1:4}})>1.0, Var({roll_{1:4}})>1.0$, resulting in the creation of the final \dataname dataset. This dataset comprises 9423 high-quality panoramas, rigorously filtered from the original collection. Building upon the \dataname dataset, we fine-tune Flux to develop a panorama LoRA~\cite{lora} for reference panorama generation.

\subsubsection{Layer Decomposition.}

 As shown in~\cref{fig:pipeline}, the reference panorama is decomposed by first identifying the scene assets and then cluster these assets in different layers according to depth. First, we employ an off-the-shelf panoptic segmentation model~\cite{jain2023oneformer} pretrained on ADE20K~\cite{ade20k} to automatically find all scene assets visible in the reference panorama. 
 A good layer decomposition requires that the layer assets share a similar depth level within layers and are distant from assets in other layers. In this sense, we assign each asset a depth value and apply K-Means to cluster these masks into different groups. 
 Given the reference panorama depth map, the depth value for each asset mask is determined by calculating the 75th percentile of the depth values within the masked region. According to the depth values, the assets are clustered into $N$ groups from layer $0$ to $N-1$ and are merged into layer masks to guide the subsequent layer completion. 

\subsubsection{Layer Completion.}
With the layer mask, we focus on completing the unseen content caused by asset occlusion. In order to synthesize background pixels instead of creating new elements, we directly utilize the above trained panorama lora and integrate it into the Flux-Fill model to accomplish domain transfer, thereby employing it as a panoramic canvas inpainter. 
Specifically, at each layer, our model takes the layer mask $M_l$, the reference panorama, and the ``\texttt{empty scene, nothing}''~\cite{zhang2024layerdiffusion} prompt as input, and output coherent content at the masked area. The inpainted panorama at layer $l$ is denoted $P_{l}$ and is used as supervision to the subsequent panoramic 3D Gaussian scene optimization. Note that, we additionally apply SAM~\cite{sam} to extend the layer mask, based on the inpainted panorama from the previous layer, to eliminate unwanted new generations from inpainting.

Moreover, to enable large-range rendering in 3D, where observers can examine scenes from varying distances, the unprocessed textures of distant assets may appear blurred as the observer approaches. Therefore, distant layers require higher resolution to preserve texture details at different viewpoints. To address this, Super Resolution (SR) module~\cite{pasd} is employed to enhance the resolution of the layered panorama from layer $0$ (background layer) to layer $N$ (reference panorama), achieving a $2\times$ upscale in resolution. SR processing significantly improves the texture quality of distant objects, maintaining their visual clarity and texture details even when observed from a closer perspective.

\subsubsection{Layer Alignment.}
Given the Layered 3D RGB Panorama $[P_{l}]_{l=0}^{N} $, we perform the depth prediction and alignment to ensure consistency in a shared space. To begin with, we apply the 360MonoDepth~\cite{rey2022360monodepth},
to first estimate the layer $N$ (reference panorama) as the reference depth $P_{depth}^{N}$. Then, to align the layer depth in 3D space, we find it infeasible to simply compute a global shift and scale as in ~\cite{luciddreamer, text2room} due to the nonlinear nature of ERP. Therefore, we leverage depth inpainting model $\mathcal{F}_{depth}$ from~\cite{liu2024infusion} to directly restore depth values based on the inpainted RGB pixels. $\mathcal{F}_{depth}$ harnesses strong generalizability from large-scale diffusion prior and synthesizes inpainted depth values at an aligned scale with the base depth. We start from reference panoramic depth $P_{depth}^{N}$ to implement step-by-step restoration from layer $N-1$ to layer $0$:
\begin{equation}
    P_{depth}^{l} = \mathcal{F}_{depth}(P_{l}, M_{l} \odot P_{depth}^{l+1}),
\end{equation}
where, in layer $l$, the inpainted panorama $P_l$ and masked depth map $M_{l}\odot P_{depth}^{l+1}$ are provided as inputs to $\mathcal{F}_{depth}$ for restoration.

\subsection{Panoramic 3D Gaussian Scene Optimization}
\label{sec:method2}

\begin{figure}[t]
  \includegraphics[width=0.9\linewidth]{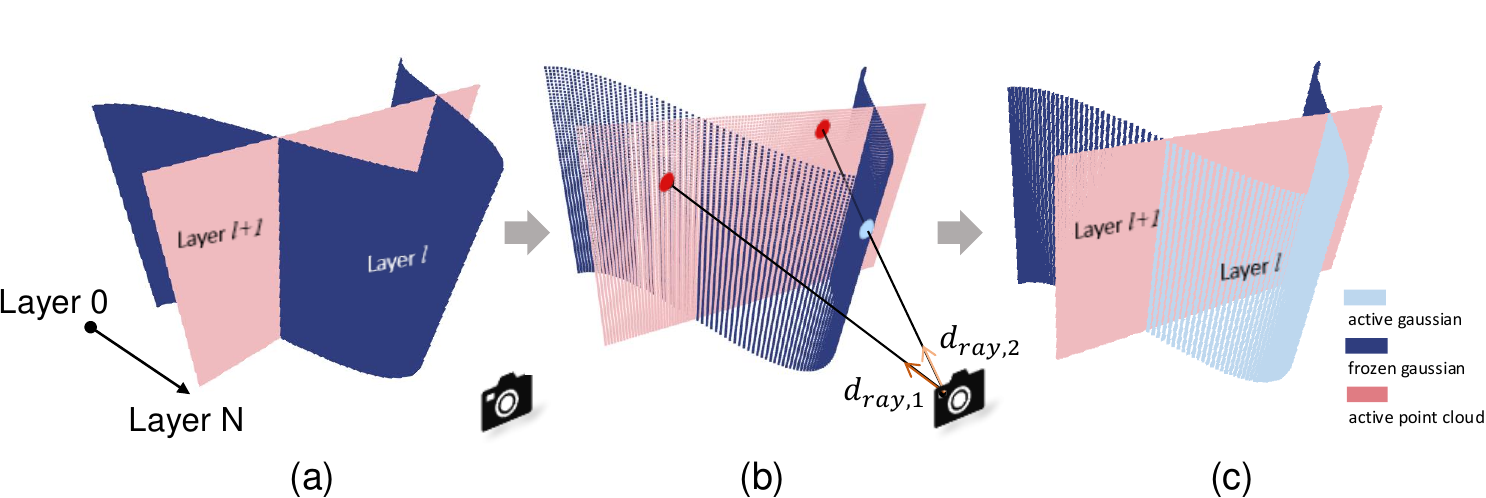}
  \caption{\small \textbf{Illustration of the Gaussian Selector.} Given the new asset point cloud, the Gaussian Selector identifies the active Gaussians for next layer's optimization. 
  }
  \label{fig:overlap_layers}
  \vspace{-2em}
\end{figure}

\subsubsection{3D Scene Initialization.}
To enable large-range 3D exploration, we lift the Layered 3D Panorama to 3D Gaussians~\cite{3dgs}, where the Gaussians are initialized from the layered 3D panoramic point clouds. Considering the intrinsic spherical structure of panorama, we can easily transform an equirectangular image $P \in \mathbb{R}^{H \times W \times 3}$ into 3D point cloud $S(\theta, \phi, P_{depth})$. Each pixel $(u,v)$ is represented as a 3D point and the angles $\theta, \phi$ are computed as $\theta = (2u/W - 1 )\pi$, $\phi = (2v/H - 1)\pi / 2$.

Then, the corresponding 3D coordinates $(X,Y,Z)$ from the depth value $P_{depth}(\theta_u, \phi_v)$ are derived as follows:
\begin{equation}
\begin{aligned}
X &= P_{depth}(\theta_u, \phi_v) \cos{\phi_v} \cos{\theta_u},\\
Y &= P_{depth}(\theta_u, \phi_v) \sin{\phi_v}, \\
Z &= P_{depth}(\theta_u, \phi_v) \cos{\phi_v} \sin{\theta_u}.
\end{aligned}
\end{equation}
Based on this transformation, we can extract the point cloud for each layer panorama to initialize 3D Gaussians.

Drastic depth changes at layout edges introduce noisy stretched outliers that would turn into artifacts during scene refinement. Therefore, we propose an outlier removal module that specifically targets stretched point removal using heuristic point cloud filtering strategies. 
As stretched points are usually sparsely distributed in space, we design the point filtering strategy based on its distance from the neighbors. First, we filter out all points with the minimum distance to neighbors over threshold $\beta_1$. Then, we eliminate points with very few neighbors. The idea is to calculate the number of neighbors of each point within a given radius and drop the points where their number of neighbors is below threshold $\beta_2$. To speed up the calculation, we map the points into 3D grids, then remove all points within grids that have less than $\beta_2$ number of neighbors.  

\subsubsection{3D Scene Refinement.}

During scene refinement, we devise two types of Gaussian training schemes for varying scene content: the \textit{base Gaussian} for reconstructing the scene background and the \textit{layer Gaussian} for optimizing scene layouts. 
Additionally, a Gaussian selector module is introduced between layer Gaussians to facilitate scene composition. 

In scene refinement, the base Gaussian model is initialized on a whole of the background point cloud, and the layer Gaussian model initiates on and optimizes the foreground assets.
In practice, we project the layer mask $\mathcal{M}_l$ onto point clouds and use the masked points to initiate Gaussians. The optimized Gaussians from previous layers are frozen to avoid unwanted modification. In this way, the scene background is optimized once in the base Gaussian to reduce unnecessary computation and conflicts of Gaussians in subsequent layers. 

We observe that the quality of the optimized scene is easily hampered by unaligned layers, and sometimes $\mathcal{F}_{depth}$ does fail to produce perfectly aligned layer depths. Gaussians at layer $l$ could span into unwanted depth levels and block assets in the subsequent layer, as illustrated in~\cref{fig:overlap_layers}(a). To handle this issue, we introduce the {Gaussian selector module} to detect these conflicted Gaussians, re-activate them from frozen, and optimize them away from the blockage. First, the selector computes the distance vector from the camera center $\mathbf{o} = (0,0,0)$ to each new point $\mathbf{p}$, as in~\cref{fig:overlap_layers}(b). The absolute distance from asset points $\mathbf{p}$ and scene Gaussians $\mathbf{g}$ to the camera is denoted as $d_\mathbf{p}$ and $d_\mathbf{g}$ respectively:$d_\mathbf{p} = || \mathbf{p} - \mathbf{o}||_2, \quad d_{\mathbf{g}} = || \mathbf{g} - \mathbf{o}||_2$. By examining all Gaussians that on the same ray with asset points but at a closer distance: $\mathbf{p} \ / \ d_\mathbf{p} = \mathbf{g} \ / \ d_\mathbf{g}, \quad d_{\mathbf{g}} < d_{\mathbf{p}},$ we mark them as active (~\cref{fig:overlap_layers}(c)). For efficient memory storage and fast look-up, we hash the distance vectors into a 3D grid. The mapping function from vector coordinates to grid indices writes: $f(\mathbf{p}) = \text{ceil}(\beta_3 \log(\textbf{p}+1))$.

\section{Experiments}
\subsection{Implementation Details} 
\label{sec:implementation_details}
In the layered panorama construction stage, we train the panorama LoRA starting from Flux~\cite{flux2023} with a batch size of 1 and learning rate of $10^{-5}$ for 100K iterations on our curated \dataname dataset. The training is done using 6 NVIDIA A100 GPUs for 3 days. For Layer Decomposition, we employ OneFormer~\cite{jain2023oneformer} to obtain the panoptic segmentation map for the reference panorama. Background categories are manually determined (i.e. sky, floor, ceiling, etc.) to filter out background components in asset masks. Generally, we cluster all asset masks into $N=3$ layers via KNN and merge all masks within each layer to form a unified layer mask. With the obtained layer mask, we integrate the above trained panorama LoRA into FLux-Fill model and combine with LaMa~\cite{suvorov2021resolutionrobust} to achieve multi-layer completion and apply 360MonoDepth~\cite{rey2022360monodepth} to predict the reference panorama depth.  In the 3D panoramic scene optimization stage, we lift the panorama RGBD into 3D point clouds. For scene initialization, we set $\beta_1$ to $0.0001$ and $\beta_2$ to $4$ based on empirical practice. These point clouds are used to initialize the base Gaussian model and the layer Gaussian model. During the scene refinement stage, we optimize the base Gaussian model for 3,000 iterations, then the layer Gaussians each for 2,000 iterations. The training objective for base and layer Gaussian is the L1 loss and D-SSIM term between the ground-truth views and the rendered views. We use a single 80G A100 GPU for reconstruction and the reconstruction time for each layer is 1.5 minutes on average for $1024\times 1024$ resolution inputs.

\subsection{Comparison Methods.}
To evaluate the performance of our approach in text-driven 3D panoramic scene generation domain, we compare with existing methods in two phases: \textbf{2D Panorama Generation} and \textbf{3D Panoramic Scene Reconstruction}. For 2D Panorama Generation, we compare the quality and creativity of 2D panorama with \textit{Text2light}~\cite{chen2022text2light} (GAN-based HDR panorama generation), \textit{Diffusion360}~\cite{feng2023diffusion360} (diffusion-based text-to-panorama generation) and \textit{Panfusion}~\cite{zhang2024taming} (dual-branch diffusion-based generation). For 3D Panoramic Scene Reconstruction, we compare with \textit{Text2Room}~\cite{text2room} (iterative indoor scene expansion with textured mesh), \textit{LucidDreamer}~\cite{luciddreamer} (single-view scene generation with 3DGS),  and \textit{Dreamscene360}~\cite{zhou2024dreamscene360} (text-guided panoramic 3DGS scene generation).

\subsection{Qualitative Comparison}
\label{sec:Qualitative}
\subsubsection{2D Panorama Generation.}
We show some qualitative comparisons with several state-of-the-art panorama generation works in~\cref{fig:pano_exp}. Text2Light~\cite{chen2022text2light} struggles to effectively interpret text prompt due to being trained on a realistic HDRI dataset based on the VQGAN structure, and the components in the generated panorama are relatively simple. The results by PanFusion~\cite{zhang2024taming} are ambiguous and low in quality. While the instances generated by Diffusion360~\cite{feng2023diffusion360} exhibit superior quality in comparison to the aforementioned methods, they lack intricate scene details and are prone to the generation of artifacts. In contrast, our method achieves the highest quality, presenting creative and reasonable generations. 
\subsubsection{3D Panoramic Scene Reconstruction.}
We present qualitative comparisons with Text2Room~\cite{text2room}, LucidDreamer~\cite{luciddreamer}, and DreamScene360~\cite{zhou2024dreamscene360} across two dimensions. \textit{First}, for full $360^\circ \times 180^\circ$ view consistency, we render multiple views from scene center point with single input image and text prompts. As shown in~\cref{fig:exp_360}, LucidDreamer~\cite{luciddreamer} and Text2room~\cite{text2room} fail to cover the full $360^\circ \times 180^\circ$ view, resulting in semantic incoherence and artifacts due to their successive inpainting-based strategy.
DreamScene360~\cite{zhou2024dreamscene360} supports a $360^\circ \times 180^\circ$ view at a single fixed viewpoint, but the quality of the generated results is relatively low. In contrast, our model excels in maintaining full $360^\circ \times 180^\circ$ view consistency while demonstrating superior content creativity. 
\textit{Second}, to evaluate novel path rendering, we design a zigzag trajectory to guide the camera's movement through the scene, with novel view renderings sampled along the trajectory for comparison. \cref{fig:exp_zigzag} shows 6 random samples from this fixed flythrough trajectory. Compared with all three methods, our model achieves a more complete 3D scene with consistent textures and a reasonable geometric structure.

\begin{figure*}[t]
	\centering
	\includegraphics[width=1.0\linewidth]{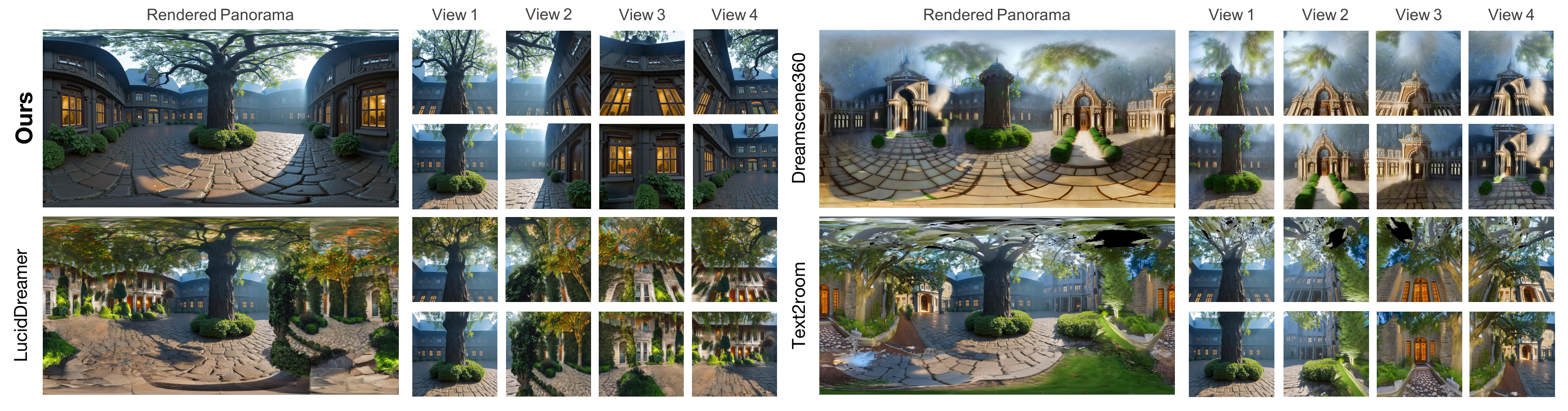}
        \setlength{\abovecaptionskip}{-3mm}
	\caption{\small
	\textbf{Qualitative comparisons in full 360°×180° Scene.} We compare the panorama and multiple views of the scene generated by four methods. ~\name~ exhibits consistent and rich details across full $360^\circ \times 180^\circ$ coverage, while other methods show obvious inconsistencies and disorganized patterns in regions that deviate from the input view.
	}
	\label{fig:exp_360}
    \vspace{-10pt}
\end{figure*}

\begin{figure*}[t]
	\centering
	\includegraphics[width=1.0\linewidth]{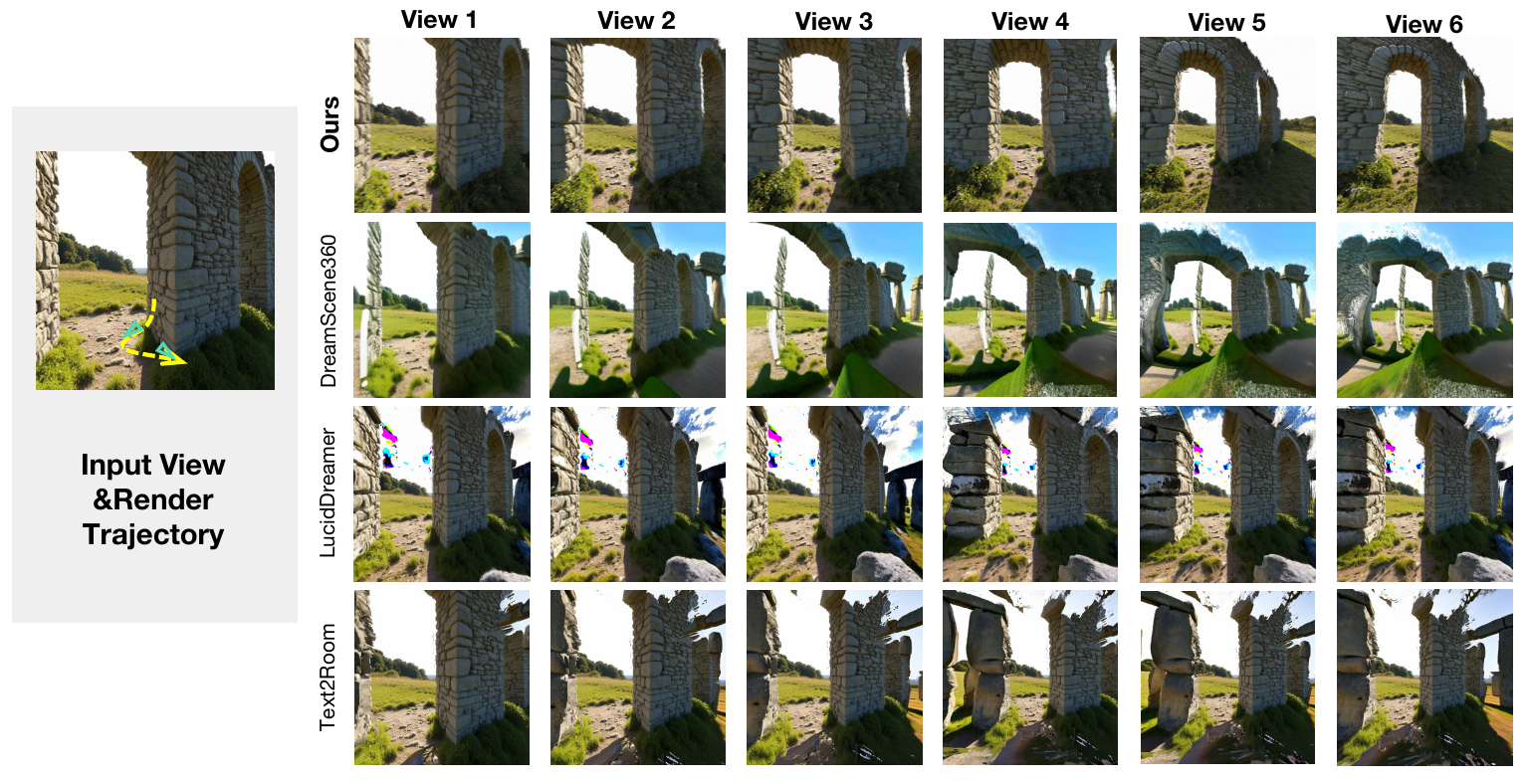}
        \setlength{\abovecaptionskip}{-3mm}
	\caption{\small
	\textbf{Qualitative comparisons in Large-range Scene Exploration.} We show the novel view renderings along a zigzag trajectory to compare the capability of large-range scene exploration. Our method is able to maintain high-quality content rendering and does not show distortion or gaps in unseen space, which shows the ability of ~\name~ to create hyper-immersive panoramic scenes.
	}
	\label{fig:exp_zigzag}
    \vspace{-10pt}
\end{figure*}

\begin{table}[t]
    \small
    \centering
    \setlength{\tabcolsep}{5pt}
    \renewcommand{\arraystretch}{0.8}
    \caption{\textbf{Quantitative comparison with SoTA methods on 2D Panorama Generation.} \textbf{Bold} indicates the best result.}
    \vspace{-6pt}
    \begin{tabular}{@{}l|cccc@{}}
    \toprule
    Method &  FID~$\downarrow$ & Aesthetic~$\uparrow$ & CLIP~$\uparrow$ & User Study (AUR)~$\uparrow$  \\ 
    \midrule
        Text2light& 286.90 & 4.57 & 18.69 & 1.34\\
        Panfusion & 283.80 & 4.78 & 21.22 &  2.38\\ 
    Diffusion360  & 274.03 & 5.07 & 21.65 & 2.52\\ 
        Ours &  \textbf{223.51} & \textbf{5.86} & \textbf{22.25} & \textbf{3.76} \\ 
        \bottomrule
    \end{tabular}
    \vspace{-12pt}
  \label{tab:metrics}
\end{table}

\begin{table*}[t]
    \small
    \centering
    \setlength{\tabcolsep}{8pt}
    \caption{\textbf{Qualitative comparison with SoTA methods on 3D Panoramic Scene.} \textbf{~\name~} achieves high-quality reconstruction and novel view synthesis while maintaining upright panoramic scene compared to other methods.}
    \renewcommand{\arraystretch}{0.8}
    \vspace{-6pt}
    \scalebox{0.9}{\begin{tabular}{@{}l|cc|ccc|cc|cc@{}}
    \toprule
    \multirow{2}{*}{Method} & \multicolumn{5}{c|}{Appearance} & \multicolumn{2}{c|}{Geometry}   & \multicolumn{2}{c}{User Study (AUR)}    \\ 
     \cmidrule(l{0.5em}r){2-10} 
         &  {NIQE~$\downarrow$} &  {BRISQUE~$\downarrow$} &  {PSNR~$\uparrow$} &  {SSIM~$\uparrow$} &  {LPIPS~$\downarrow$}  & Pitch-Mean~$\downarrow$  & Pitch-Var~$\downarrow$ &\begin{tabular}[c]{@{}c@{}}$360^\circ\times 180^\circ$~$\uparrow$ 
    \end{tabular}  & \begin{tabular}[c]{@{}c@{}}Free-path~$\uparrow$ \end{tabular}\\
    \midrule
        Text2room~\cite{text2room} & {5.231} & 46.127 & 30.126 & 0.882	& 0.038 & 2.029 & 1.724 & 1.69 & 2.31\\
        LucidDreamer~\cite{luciddreamer} & 5.822 & 52.102 &  {36.108} &  {0.954} &  {0.026} &2.813 & 2.189 & 1.81  & {1.31} \\ 
        DreamScene360~\cite{zhou2024dreamscene360}  &  {5.051}	&  {39.891} & 30.056 & 0.958 & 0.062 & 1.328 & 2.018 &  {2.86}   &  {2.77} \\ 
        Ours & \textbf{4.023} & \textbf{38.287} & \textbf{42.057} & \textbf{0.986} & \textbf{0.015}	& \textbf{0.732} & \textbf{0.032} & \textbf{3.64} & \textbf{3.61} \\ 
        \bottomrule
    \end{tabular}}
    \vspace{-8pt}

  \label{tab:3drecon_metrics}
\end{table*}

\subsection{Quantitative Comparison}
\subsubsection{2D Panorama Generation.}
We adopt three metrics for quantitative comparisons: 1) \textit{FID}~\cite{FID} evaluates both fidelity and diversity; 2) \textit{Aesthetic}~\cite{aesthetic} evaluates the aesthetics of panorama; 3) \textit{CLIP}~\cite{Clipscore} measures the compatibility of results with input prompts. 
Moreover, a user study is also conducted to further evaluate the quality of panoramas, where we project 4 views at a fixed FOV ($90^\circ$) to the user for sorting. Here, we report the average {Average User Ranking (AUR)}~\cite{controlnet}, which is computed based on an integrated assessment of coherence, plausibility, aesthetics, and compatibility dimensions. As shown in~\cref{tab:metrics}, our method achieves the best scores among all quantitative metrics and human evaluation, demonstrating its fidelity, alignment with text and overall consistency.

\subsubsection{3D Panoramic Scene Reconstruction.}
Following \cite{zhou2024dreamscene360}, we adopt non-reference image quality assessment metrics, \textit{NIQE}~\cite{niqe} and \textit{BRISQUE}~\cite{brisque}, to evaluate novel view quality along scene navigation paths. We also follow \cite{zhou2024dreamscene360} to measure the rendering quality with \textit{PSNR}, \textit{SSIM} and \textit{LPIPS}~\cite{lpips}. In terms of geometry evaluation, we render 4 orthogonal views ($90^\circ$ FOV, $0^\circ$ elevation and \{ $0^\circ, 90^\circ, 180^\circ, 270^\circ$ \} azimuths) and predict the \textit{Pitch-Mean} and \textit{Pitch-Var} (mean and variance of the elevation angles) with~\cite{GeoCalib} to evaluate whether the scenes are upright. As shown in~\cref{tab:3drecon_metrics}, our method surpasses the existing methods in both novel view quality metrics (NIQE and BRISQUE), 3D reconstruction metrics (PSNR, SSIM, LPIPS) and while ensuring the upright panoramic scene. 
Furthermore, we conduct another user study for 3D panoramic scene evaluation from two aspects: 1) $360^\circ \times 180^\circ$ view consistency and 2) novel path rendering quality. For the first aspect, we render 60 frames to cover 360-degree view at the 0-degree and 45-degree elevation respectively for evaluation. For the second aspect, we use the same trajectory as in~\cref{fig:exp_zigzag} to render navigation videos for evaluation.
We invite 52 users including graduate students that expertise in 3D and average users to rank the 40 results from 4 methods.
The average ranking is shown in~\cref{tab:3drecon_metrics}. Our~\name~achieves the best performance in both $360^\circ \times 180^\circ$ view consistency and novel path rendering quality among all four approaches.

\begin{figure}[t]
	\centering
	\includegraphics[width=1.0\linewidth]{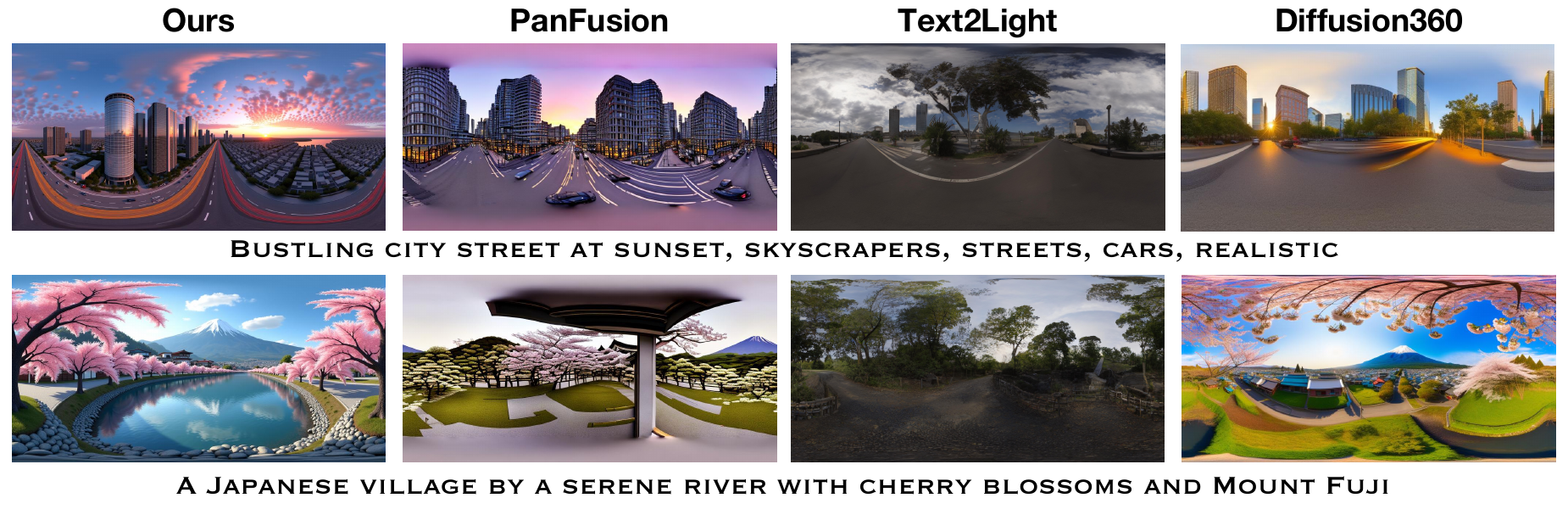}
        \setlength{\abovecaptionskip}{-3mm}
	\caption{\small
	\textbf{Qualitative comparisons in Panorama Generation.} \name~demonstrates superior capability in generating high-quality outputs with precise alignment to text prompt, outperforming other methods in fidelity and input adherence.
	}
	\label{fig:pano_exp}
    \vspace{-2em}
\end{figure}

\subsection{Analysis and Ablative Study}
In this section and supp., we show the analysis and ablation on the Gaussian Selector (\cref{sec:gaussian_selector}), Multi-layer design (single vs. multi;~\cref{sec:off_center}), Layer Gaussians representation (supp.),  layer inpainting (supp.) and 3DGS optimization efficiency (supp.).

\subsubsection{Ablation on Gaussian Selector.}
\label{sec:gaussian_selector}
Our Gaussian selector is proposed to select the part of Gaussians that appears in the front of newly added scene assets. By selecting these Gaussians and re-activating them in the optimization, the model achieves accurate appearance and geometry at the current layer. As shown in~\cref{fig:ablate_gaussian_selector}, the leftmost column is the scene Gaussians at layer 0. When adding the building assets at the first layer, the sky Gaussians from the previous layer partially block the building assets (right column). After using the Gaussian selector to select and optimize the sky Gaussians, these Gaussians learn to either be translucent and pruned for low opacity or move to be a part of the building assets. Therefore in the middle column, we observe a consistent scene with no obvious blockage of the new building assets thanks to the Gaussian Selector.

\begin{figure}
	\centering
	\includegraphics[width=0.95\linewidth]{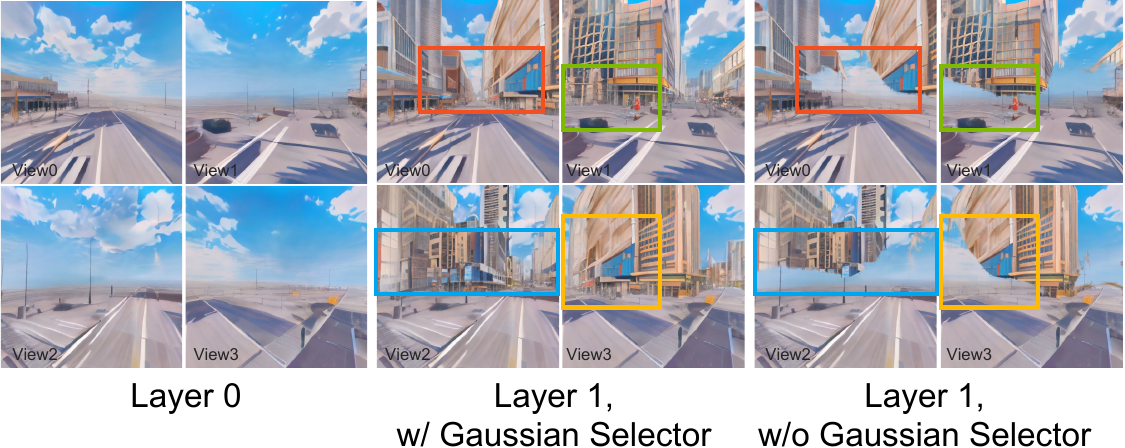}
        \setlength{\abovecaptionskip}{0.01mm}
	\caption{\small
    \textbf{Ablation on the Gaussian Selector.} 
    With the Gaussian Selector, the merged Gaussians are optimized to faithfully reconstruct the ground-truth panorama views.}
	\label{fig:ablate_gaussian_selector}
        \vspace{-2em}
\end{figure}

\begin{figure}
	\centering
    \vspace{-2pt}
	\includegraphics[width=0.92\linewidth]{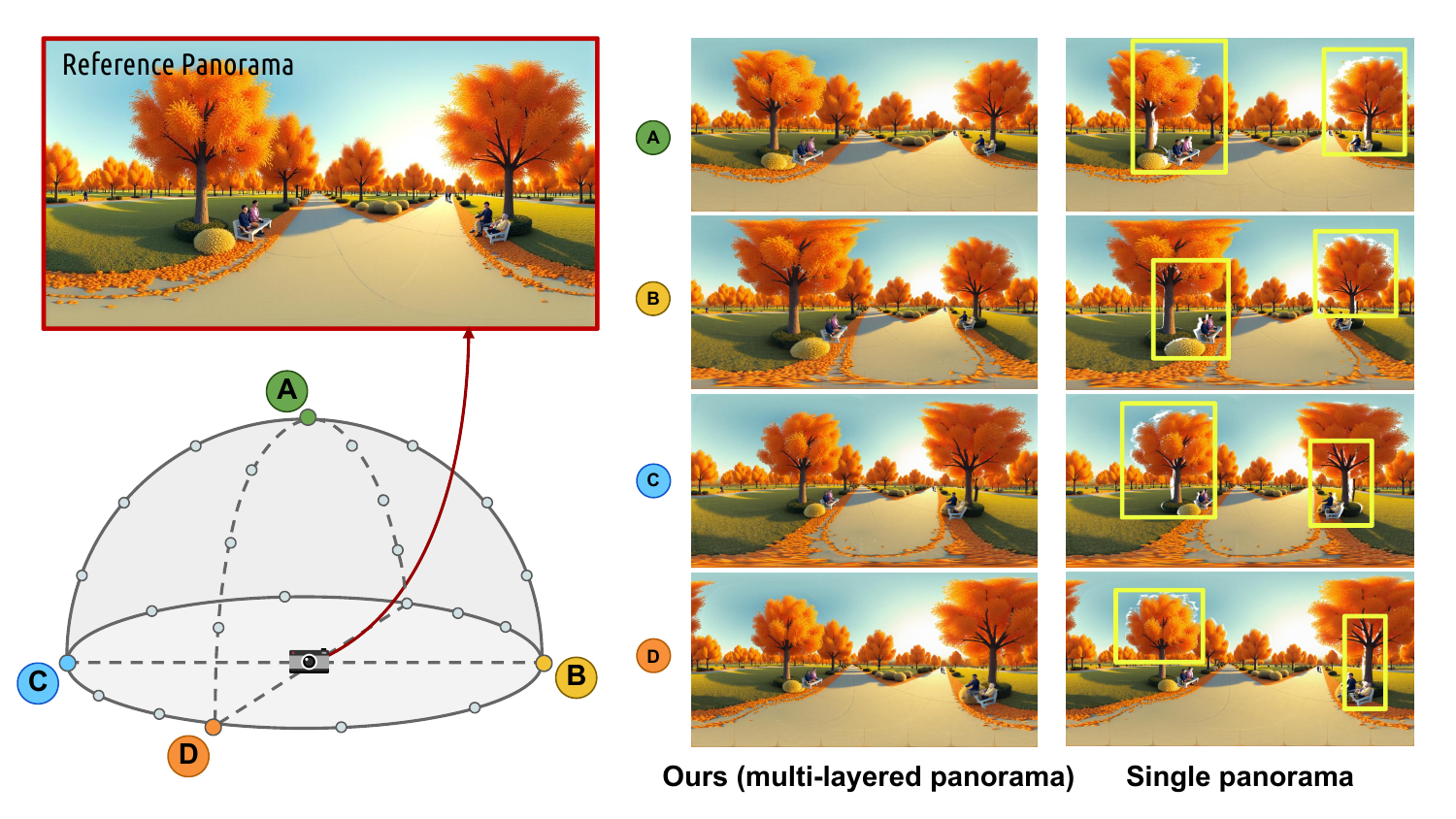}
        \setlength{\abovecaptionskip}{-0.1mm}
	\caption{\small
    \textbf{Analysis on Panorama Rendering at Off-center Viewpoints.} 
    Compared with the single-layer variant,~\name~render $360^\circ\times 180^\circ$ consistent panorama at various off-center viewpoints without any holes or gaps from occlusion.}
	\label{fig:ablate_abc}
        \vspace{-2em}
\end{figure}

\subsubsection{Analysis on Panorama Renderings at Off-center Viewpoints.}
\label{sec:off_center}
In~\cref{fig:ablate_abc}, we demonstrate that~\name~is robust to render consistent panorama images at various locations besides the original camera location in the center. We sample four camera locations on circular trajectories on the hemisphere centered at the origin and render 24 views at ($-45^\circ, 0^\circ, 45^\circ$) elevation to compose new panorama images. By evaluating panorama renderings at new viewpoints, we show that our generated panoramic scene is $360^\circ\times 180^\circ$ consistent and enclosed, robust to various viewpoints at any angle. Compared to the single-layered 3D panorama, our multi-layered 3D panorama exhibits no gaps or holes from the scene occlusion, demonstrating our capability for larger-range, complex 3D exploration in the generated scenes.

\section{Conclusion}
In this paper, we propose~\name, a novel framework that generates hyper-immersive panoramic scene from a single text prompt. Our key contributions are two-fold. First, we propose the text-guided anchor view synthesis pipeline to generate detailed and consistent reference panorama. Second, we pioneer the Layered 3D Panorama representation to show complex scene hierarchies at multiple depth layers, and lift it to Gaussians to enable large-range 3D exploration.
Extensive experiments show the effectiveness of~\name~in generating $360^\circ \times 180^\circ$ consistent panorama at various viewpoints and enabling immersive roaming in 3D space. We believe that~\name~holds promise to advance high-quality, explorable 3D scene creation in both academia and industry.

\noindent\textbf{Limitations and Future Works.}
\name~leverages good pre-trained prior to construct panoramic 3D scene, i.e., panoramic depth prior for 3D lifting. Therefore, the created scene might contain artifacts from inaccurate depth estimation.
With advancements in more robust panorama depth estimation, we hope to create high-quality panoramic 3D scenes with finer asset geometry.

{
\bibliography{main}
\bibliographystyle{ACM-Reference-Format}
}

\begin{figure*}
    \centering
    \vspace{-14pt}
    \includegraphics[height=\textheight]{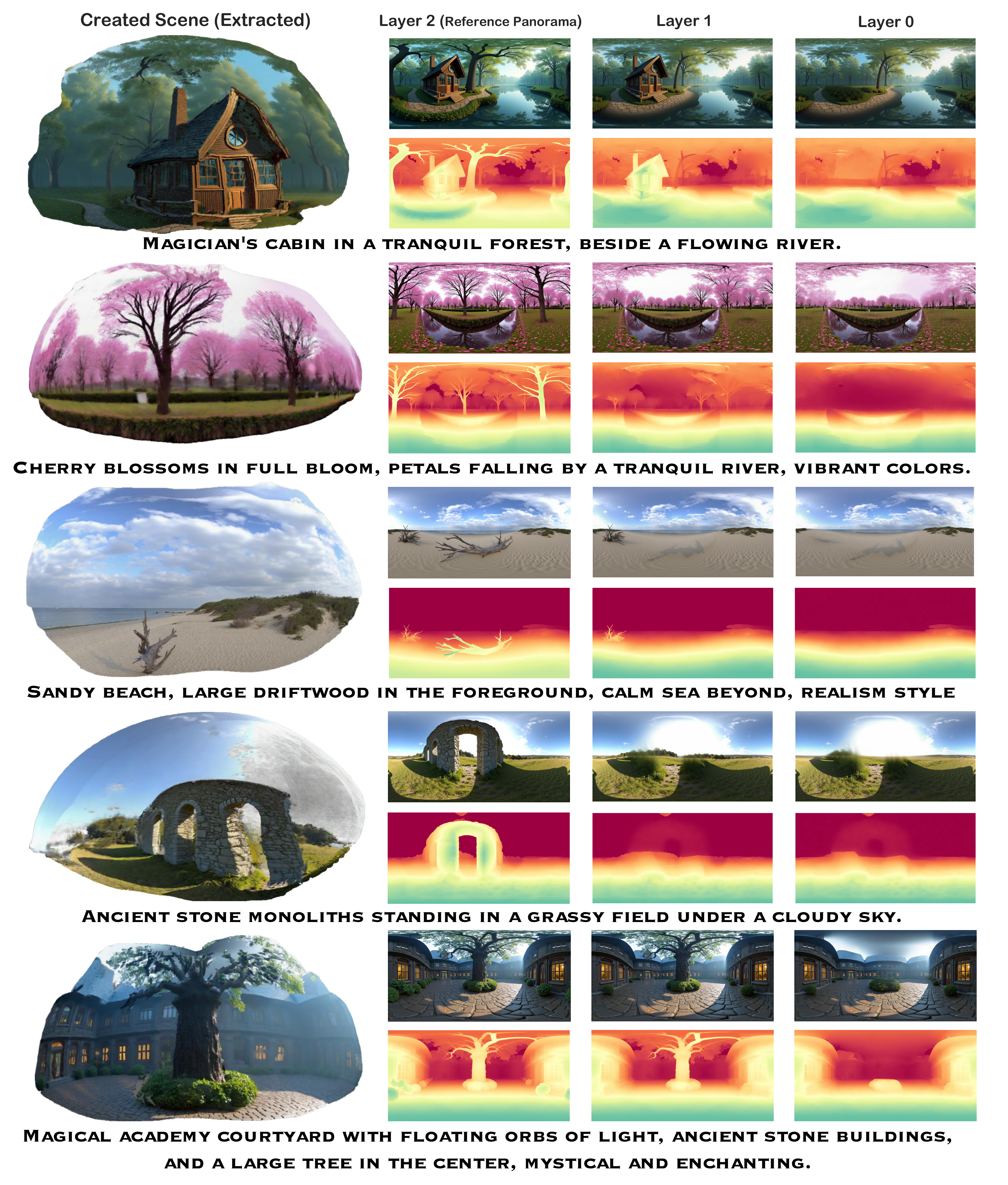}
    \vspace{-2em}
    \caption{\textbf{Additional results of ~\name~on Diverse Generation.}~\name~generates various hyper-immersive scene with consistent and rich details across full $360^\circ \times 180^\circ$ coverage.}
    \label{fig:fig_only}
\end{figure*}

\begin{figure*}[h]
	\centering
	\includegraphics[width=0.95\linewidth]{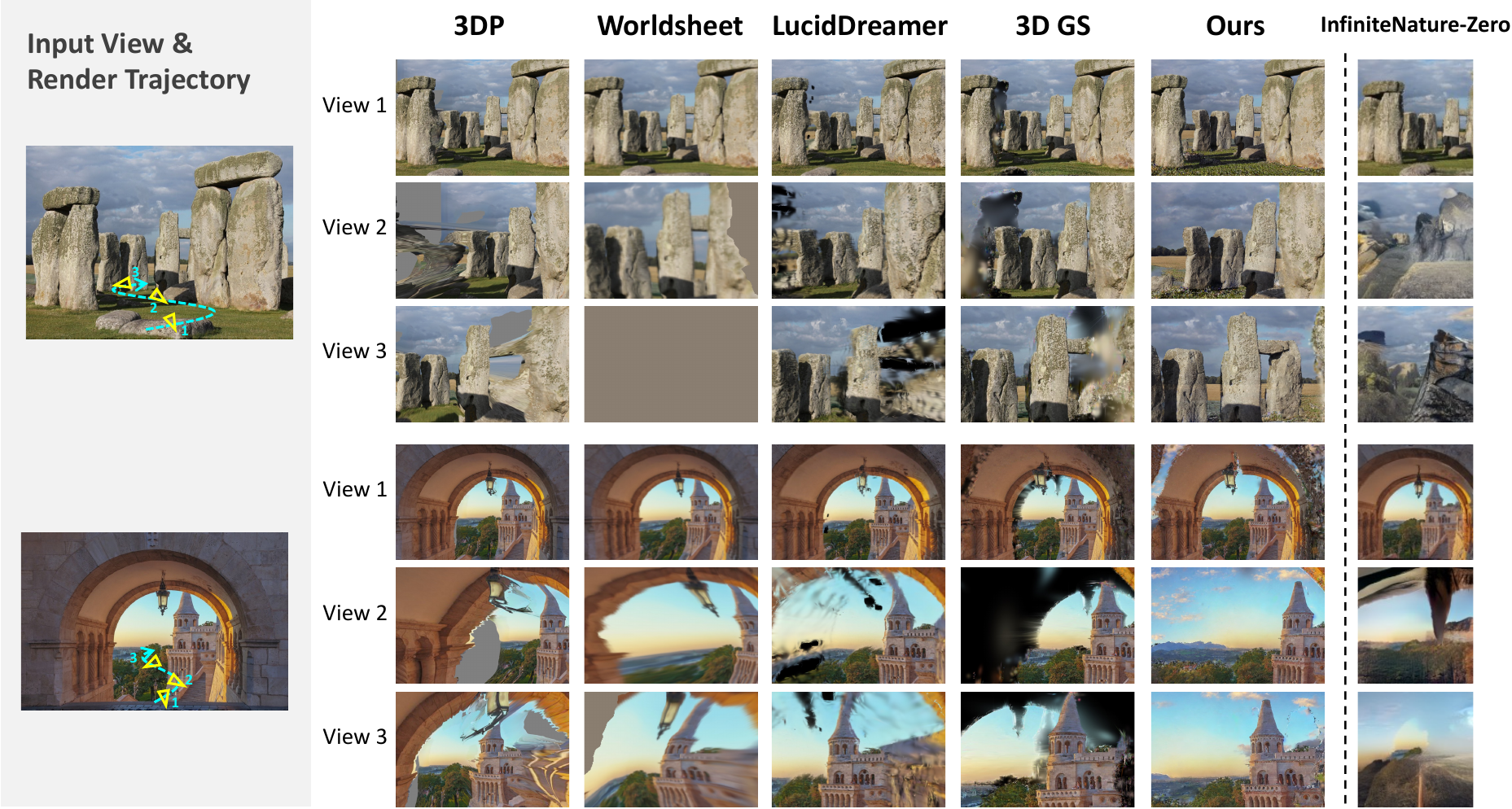}
        \setlength{\abovecaptionskip}{-0.1mm}
	\caption{\small
    \textbf{Analysis on Layer Gaussians Representation} with 3DP~\cite{3dp}, Worldsheet~\cite{worldsheet}, LucidDreamer~\cite{luciddreamer}, and Single-view 3D GS~\cite{3dgs} on novel view renderings along a zigzag trajectory. InfiniteNature-Zero~\cite{infinitezero} are shown with three random views from its fixed trajectory.
	}
	\label{fig:single_exp}
        \vspace{-2pt}
\end{figure*}

\begin{figure*}[h]
	\centering
	\includegraphics[width=0.95\linewidth]{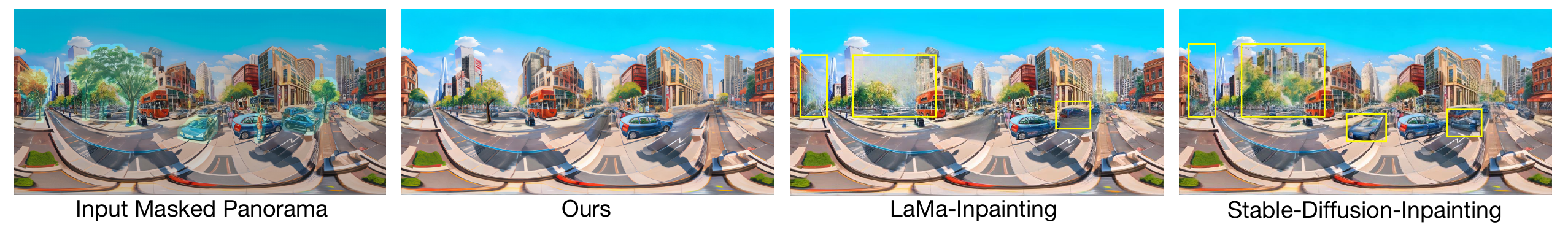}
        \setlength{\abovecaptionskip}{-0.1mm}
	\caption{\small
    \textbf{Analysis on the Layer Completion Inpainting.} We present the panorama inpainting results for three methods guided by the same text prompt: ``\texttt{empty scene, nothing}''~\cite{zhang2024layerdiffusion}. Our model effectively handles complex scenarios, delivering clear results with consistent and coherent structures.
	}
	\label{fig:ablation_inpaint}
        \vspace{-1em}
\end{figure*}

\section{Additional Experiment Details}
\subsection{More Evaluation Details on 2D Panorama Generation.}

We use various metrics to evaluate the coherence, fidelity, diversity, aesthetic and compatibility of generated panoramas with the input prompt.

\noindent• \textbf{(Fidelity \& Diversity)  FID}~\cite{FID}: 
Fréchet Inception Distance(FID) is employed to assess both fidelity and diversity. We calculate FID between panoramas from Matterport3D~\cite{Matterport3D} and the generated panoramas.

\noindent• \textbf{(Aesthetic)  Aesthetic}~\cite{aesthetic}: 
For each panorama, we randomly project 20 views at a fixed FOV ($90^\circ$) with resolution $512 \times 512$ and calculate their average aesthetic scores.

\noindent• \textbf{(Compatibility)  CLIP}~\cite{Clipscore}: Compatibility with the input prompt is evaluated using the mean of CLIP scores, mirroring the approach for aesthetic evaluation.

\noindent • \textbf{(Coherence)  Intra-Style}~\cite{intra-style}: 
To assess coherence, we introduce Intra-Style, computed as the average Style Loss between pairs of window images from the same panorama. We begin by resizing the panorama to $512 \times 1024$, then crop it into 4 windows with a stride of 256. The final image is formed by seamlessly connecting the panorama's tail and head. Each window is $512 \times 512$, and we compute the average Style Loss across the 6 combinations of these cropped views.

Table~\ref{tab:sup_metrics} presents a quantitative comparison among the methods. \textbf{Bold} indicates the best result, and \underline{underline} indicates the second-best result. Our method achieve the optimal scores in FID, Aesthetic and CLIP metrics, which indicates the high quality in creativity, fidelity, and compatibility with input prompts of our method. The results of Intra-Style demonstrate that our method achieves global coherence across the image, maintaining a consistent overall style. Although Text2light~\cite{chen2022text2light} has a smaller Intra-Style score, this is due to its tendency to generate monotonous panoramas with extensive uniform color block backgrounds. Moreover, the generated contents are largely unrelated to the guidance provided by the input prompts. Consequently, the metric of Intra-Style for Text2light has no comparison significance.

\section{Additional Analysis and Ablative Study}

\subsection{Analysis on 3DGS optimization efficiency.}
For time efficiency, as we mentioned in Sec. 4.1, we use a single 80G A100 GPU for 3DGS optimization and the optimization time for each layer is 1.5 minutes on average for $1024\times 1024$ resolution inputs. 
For memory efficiency, if we use a pixel-aligned 3DGS for optimization, we would easily encounter OOM as the layers increase. Here we have two steps to reduce the memory cost. First, as we described in methods section, we select new assets at each layer and only optimize the new assets combined with active Gaussians at each layer. In this way, our model does not introduce additional Gaussians to represent the same asset. Second, although we use layer mask to select point clouds in a pixel-aligned manner, but we downsample the point cloud to be under $N_{max}$ points at each layer before 3DGS initialization to not exceed the GPU memory, and remain a small size for visualization and rendering.  Empirically, we set $N_{max}$ to 2,000,000. We also show a breakdown of GPU memory usage at each layer for a random case in Table~\ref{tab:3dgs_optim}. The maximum memory usually does not exceed 3000 MB for all cases.

\begin{table}[t]
    \small
    \centering
    \caption{\textbf{Quantitative comparison with SoTA methods.} \textbf{Bold} indicates the best result, and \underline{underline} indicates the second-best result.}
    \vspace{-6pt}
    \begin{tabular}{@{}l|cccc@{}}
    \toprule
    Method &   FID~$\downarrow$ & Aesthetic~$\uparrow$ & CLIP~$\uparrow$ & Intra-Style~$\downarrow$ \\ 
    \midrule
        Text2light  & 286.90 & 4.57 & 18.69  & \textbf{0.31}\\
        Panfusion & 283.80 & 4.78 & 21.22  & 18.66\\ 
    Diffusion360  & \underline{274.03} & \underline{5.07} & \underline{21.65} & 3.70 \\ 
        Ours & \textbf{223.51} & \textbf{5.86} & \textbf{22.25}  & \underline{1.63} \\ 
        \bottomrule
    \end{tabular}
    \vspace{-8pt}

  \label{tab:sup_metrics}
\end{table}

\begin{table}[t]
    \small
    \centering
    \caption{\textbf{Memory usage at each layer in Layer Gaussians Optimization.} Our optimization strategy ensures that memory consumption remains at a low level. }
    \vspace{-6pt}
    \begin{tabular}{@{}l|cccc@{}}
    \toprule
    Layer ID &   Layer 0 & Layer 1 & Layer 2 & Layer 3 \\ 
    \midrule
        Memory (MB)  & 1997.04 & 2079.75 & 2280.99  & 2507.61 \\
        Newly Added GS & 1702242 & 129215 & 157183  & 102450\\ 
        \bottomrule
    \end{tabular}
    \vspace{-8pt}

  \label{tab:3dgs_optim}
\end{table}

\subsection{Analysis on Layer Gaussians Representation.}
In the main paper, we validate the effectiveness of the Layer Gaussians representation in addressing occlusion for hyper-immersive panoramic scene generation, through experiments on full $360^\circ \times 180^\circ$ view consistency and large-scale exploratory trajectory rendering capability. 
Building on this, we extend our discussion to explore the application of this representation in single-image to scene task. We show the qualitative comparisons with 3DP~\cite{3dp}, Worldsheet~\cite{worldsheet}, InfiniteNature-Zero~\cite{infinitezero}, LucidDreamer~\cite{luciddreamer} and 3D GS~\cite{3dgs} in ~\cref{fig:single_exp}. The camera moves along a zigzag trajectory into the scene, and the novel view renderings are sampled along the trajectory for comparison among all methods. For InfiniteNature-Zero~\cite{infinitezero}, we showed 3 random samples from its fixed fly-through trajectory. Compared to all five methods, our model achieves more complete 3D scene with consistent texture and accurate geometry in both occluded and non-occluded space, demonstrating our ability of high-quality image-conditioned 3D scene creation.

\subsection{Analysis on Layer Completion Inpainting.}
We discuss the effectiveness of our panorama inpainter in layer completion. We compare the inpainting results among three approaches: LaMa~\cite{suvorov2021resolutionrobust}, stable diffusion inpainting model, and our proposed inpainter. As illustrated in~\cref{fig:ablation_inpaint}, LaMa produces inconsistent texture and blurry artifacts at large-scale inpainting. Pure stable diffusion tends to produce distorted new elements due to the domain gap between perspective and panoramic images. In contrast, thanks to the panoramic lora and the introduced controllable generation strength, our module delivers clean inpainting results with coherent and plausible structures in the masked regions.

\end{document}